\pdfoutput=1

\documentclass[11pt]{article}

\usepackage{naacl2021}
\usepackage{booktabs}
\usepackage{graphicx}

\usepackage{times}
\usepackage{latexsym}

\usepackage[T1]{fontenc}

\usepackage[utf8]{inputenc}

\usepackage{microtype}
\usepackage{paralist}

\title{Adapting Coreference Resolution for Processing Violent Death Narratives}

\author{Ankith Uppunda, Susan D. Cochran, Jacob G. Foster \\ {\bf  Alina Arseniev-Koehler, Vickie M. Mays,  Kai-Wei Chang\thanks{ \ \  \texttt{kwchang@cs.ucla.edu}}} \\
  University of California, Los Angeles }
\begin{document}
\maketitle

\begin{abstract}
Coreference resolution is an important component in analyzing narrative text from administrative data (e.g., clinical or police sources). 
However, existing coreference models trained on general language corpora suffer from poor transferability due to domain gaps, especially when they are applied to gender-inclusive data with lesbian, gay, bisexual, and transgender (LGBT) individuals. In this paper, we analyzed the challenges of coreference resolution in an exemplary form of administrative text written in English: violent death narratives from the USA's  Centers for Disease Control's (CDC) National Violent Death Reporting System.
We developed a set of data augmentation rules to improve model performance using a probabilistic data programming framework. 
Experiments on narratives from an administrative database, as well as existing gender-inclusive coreference datasets, demonstrate the effectiveness of data augmentation in training coreference models that can better handle text data about LGBT individuals.  

\end{abstract}

\section{Introduction}

Coreference resolution~\cite{soon2001machine,ng2002improving} is the task of identifying denotative phrases in text that refer to the same entity. 
It is an essential component in Natural Language Processing (NLP). In real world applications of NLP, coreference resolution is crucial for analysts to 
extract structured information from text data. Like all components of NLP, it is important that coreference resolution is robust and accurate, as applications of NLP may inform policy-making and other decisions. This is especially true when coreference systems are applied to administrative data, since results may inform policy-making decisions.

In this paper, we describe an approach to adapting a coreference model to process  narrative text from an important administrative database written in English: the National Violent Death Reporting System (NVDRS), maintained by the Centers for Disease Control (CDC) in the USA. Violent death narratives document murders, suicides, murder-suicides, and other violent deaths. These narratives are complex, containing information on one or more persons; some individuals are victims, others are partners (heterosexual or same-sex), family members, witnesses and law enforcement.
Specifically, we apply the End-to-End Coreference Resolution (E2E-Coref) system~\cite{lee2017end-to-end,lee2018higher-order}, which 
has achieved high performance on the OntoNotes 5.0~\cite{hovy2006ontonotes} corpus.
We observe that when a model trained on OntoNotes is applied to violent death narratives, the performance drops significantly for the following reasons. 

First, despite the fact that OntoNotes contains multiple genres\footnote{OntoNotes contains news sources, broadcasts, talk shows, bible and others. It consists of mostly  news-related documents.}, it does not include administrative data. Administrative text data is terse and contains an abundance of domain-specific jargon. 
Because of the gap between training and administrative data, models trained on  OntoNotes are poorly equipped to handle administrative data that are heavily skewed in vocabulary, structure, and style, such as violent death narratives.

Second, approximately 5\% of the victims in the NVDRS are lesbian, gay, bisexual, or transgender (LGBT). This is a vulnerable
population; for example, existing data show LGB youth are 5 times more likely to attempt suicide than heterosexual youth \cite{pmid32453408} and are more likely to be bullied prior to suicide\footnote{https://www.thetrevorproject.org/resources/preventing-suicide/facts-about-suicide/}. It is essential that data-analytic models work well with these hard to identify but highly vulnerable populations; 
indeed correctly processing text data is an important step in revealing the true level of elevated risk for LGBT populations. This remains challenging because of limitations of existing coreference systems.
Close relationship partners provide a marker of sexual orientations and can be used \cite{LIRA2019172, REAM2020470} by social scientists to identify relevant information in LGBT deaths.
However, OntoNotes is heavily skewed towards male entities~\cite{zhao2018gender} and E2E-Coref relies heavily on gender when deciphering context~\cite{cao2020toward}. 
Consequently, E2E-Coref has a trouble dealing with narratives involving LGBT individuals where gender referents do not follow the modal pattern.

\begin{figure}[t]
\scriptsize
\noindent\fbox{
\parbox{0.95\linewidth}{%
        \raggedright
         \colorbox{green}{primary\_victim} is \colorbox{green}{a 50 year old male}. ... \colorbox{yellow}{primary\_victim's partner} states that \color{red}\colorbox{green}{he }\color{black} and \colorbox{green}{primary\_victim} had been living together for three years. ...
    }%
}
\caption{A snippet of a violent death narrative. Highlighted is what the e2e-coref model clusters, and the colored text shows what the e2e-coref model misses.}
\label{fig:example}
\vspace{-10pt}
\end{figure}

 Figure \ref{fig:example} illustrates a scenario where coreference systems struggle. The model mislabels the pronoun ``he'' and this error will propagate to downstream analysis. Specifically, the model takes the context and resolves the coreference based on gender; it makes a mistake partially due to an incorrect presumption of the sexual orientation of the 50 year old male victim. 
 



To study coreference resolution on violent death narratives (VDN), we created a new corpus that draws on a subset of cases from NVDRS where CDC has reported the sex of both victims and their partners.  
 We assigned ground truth labels using experienced annotators trained by social scientists in public health.\footnote{All annotators have signed the release form for accessing the NVDRS data.}
 
 


To bridge the domain gap, we further adapted E2E-coref by using a weakly supervised data creation method empowered by 
the Snorkel toolkit~\cite{ratner2017snorkel}. This toolkit is often used to apply a set of rules to augment data by probabilistic programming.
Inspired by Snorkel, we designed a set of rules to 1) bridge the vocabulary difference between the source and target domains and 2)
to mitigate data bias by augmenting data with samples from a more diverse population. Because labeling public health data requires arduous human labor, data augmentation provide a promising method to enlarge datasets while covering a broader range of scenarios.

We verified our adaptation approach on both the in-house VDN dataset
as well as two publicly available English datasets, GICoref~\cite{cao2020toward} and MAP~\cite{cao2020toward}.
We then measured the performance of our approach on documents heavily skewed toward LGBT individuals and on documents in which gendered terms were swapped with non-gendered ones (pronouns, names, etc.).
On all datasets, we achieved an improvement. For LGBT specific datasets, we see much larger improvements, highlighting how poor the OntoNotes model performed on these underrepresented populations before. 
Models trained on the new data prove more applicable in that domain. 
\emph{Our experiments underscore the need for a modifiable tool to train specialized coreference resolution models across a variety of specific domains and use-cases.}

\section{Related Work}
Researchers have shown coreference systems exhibit gender bias and resolve pronouns by relying heavily on gender information~\cite{cao2020toward, zhao2018gender, rudinger2018gender,webster2018mind,zhao2019gender}. In particular, \citet{cao2020toward} collected a gender-inclusive coreference dataset and evaluated how state of the art coreference models performed against them. 



As NLP systems are deployed in social science, policy making, government, and industry, it is imperative to keep inclusivity in mind when working with models that perform downstream tasks with text data.
For example, Named Entity Recognition (NER) was used in processing Portuguese police reports to extract people, location, organization, and time from the set of documents \cite{carnaz2019review}. These authors noted the need for a better training corpus with more NER entities. 
Other NLP models face challenges in domain-adaptation like the one demonstrated in this paper. One example from the biomedical field is BioBERT \cite{Lee_2019}, in which the authors achieved better results on biomedical text mining tasks by pretraining BERT on a set of biomedical documents. Likewise, even when evaluating a model on a general set, \citet{10.1145/3366424.3383559}  showed that many general-domain datasets include as much bias as datasets designed to be toxic and biased. 
All these cases required re-evaluation of the corpus used to train the model. 
This underscores the need for methodology that can evaluate, debias, and increase the amount of data used. 

\section{Annotating Violent Death Narratives}
\label{sec:data}

We first applied for and were given access to the CDC’s National Violent Death Reporting-system’s (NVDRS) Restricted Access Database. 
From this, we sampled a total 115 of violent death cases\footnote{Homicides and Suicides} each over 200 words in length. 
In these 115 cases, we had a total of 6,134 coreference links and 44,074 tokens, with a vocabulary size of 3,653. 
Each case had information about the victim, the victim’s partner, and the type of death. 
We randomly sampled 30 cases from three strata: 1) the victim is male and the partner is female, 
2) the victim is female and the partner is male, and 3) it was an LGBT case. 
We also included 25 cases that were particularly challenging for the general E2E model.
The cases used were spell-checked and cleaned thoroughly.

\begin{figure*}[t]

\scriptsize
\centering
\noindent\fbox{
\parbox{0.95\linewidth}{%
        \raggedright
        \small
          \colorbox{green}{John Smith → J. Smith} went to the store. \colorbox{yellow}{He→Zie} wanted to buy apples, bananas, and strawberries. \colorbox{yellow}{His→Zir} girlfriend came with \colorbox{red}{him→him}, and she wanted to buy peaches and oranges. 
    }%
}
\vspace{-8pt}
\caption{
The proposed rules for GI data applied to a sample paragraph. } 
\vspace{-10pt}
\label{fig:example2}
\end{figure*}

To obtain gold-standard labels, we tasked a team of three annotators\footnote{All annotators signed the release form for accessing the NVDRS data.} to label the coreference ground truth, under the guidance of senior experts in suicide and public health.
Annotators were told that every expression referring to a specific person or group was to be placed into that person’s or group’s cluster.
From there, we resolved the three label sets into one by a majority voting method -- if two out of three annotators put the phrase in a cluster, we assigned it to that cluster. Two of the annotators had previous experience with coding the NVDRS narratives for other tasks, while one was inexperienced. Agreement was typically unanimous. 

\paragraph{Reproducibility}
  To get access to the NVDRS, Users must apply for access and follow a data management agreement executed directly with CDC. We cannot release VDN or the annotations but we will provide the augmentation code and instructions on how reproducing the experiments. To allow reproduction of our approaches on data without access-restriction, we perform evaluations on MAP and GICoref which are readily available.
\section{Weakly-Supervised Data Augmentation for Domain Adaption}
\label{sec:method}

Our next step was to build a pipeline for adapting E2E-Coref to resolve coreference on VDN.  
The key component of this pipeline is the Snorkel toolkit and its capacity to design rules that programmatically label, augment, and slice data. 
We looked to adapt E2E-Coref to process domain-specific data by creating a set of augmentation rules that would improve training data performance. Our rules can generate augmented data with diverse genders and then challenged our model to predict the coreference clusters. 


\paragraph{Data Augmentation by Rules} With Snorkel, we assessed the weakness of the current coreference model systems. These experiments helped us to develop effective augmentation rules to create training data that mimics challenging data to guide the model going forward. 
Specifically, we split data into groups and evaluated our model on split data. 
In the case of VDN, we split a larger set of data into two groups (LGBT and non-LGBT) and gauged model performance on both groups. 
We then isolated specific groups of data that posed a problem and came up with sets of augmentation rules that can be used to generate difficult training data from easier training cases. For example, in our case, we sought to augment documents that contained more precisely defined gender into cases with vaguer language regarding gender often seen in gender-inclusive documents and LGBT violent death narratives. This was seen in each rule’s effort to strip gender from key phrases, leaving it more ambiguous to the model. For example, our model struggles when terms like ‘partner’ are used to describe relationships. To address this, 
we introduced a rule where gendered relationship terms like ’girlfriend’ in one cluster were replaced by non-gendered terms like ’partner’.
In this manner, our model was forced to train against these examples. Often, the model performance improved when training against these augmented examples.

\section{Experiments and Results}

We conducted experiments to analyze E2E-Coref on VDN and verified the effectiveness of the data augmentation method. 
We used the following corpora\footnote{VDN must be obtained directly from CDC. We also conducted experiments on publicly available datasets.}. 
\begin{itemize}
\item \textbf{OntoNotes} We used the English portion of version 5.0. It contains roughly 1.6M words. 
\item \textbf{VDN} The annotated violent death narratives described in Sec. \ref{sec:data}. The corpus is annotated by domain experts and used as the test set for measuring model performance. We split VDN into train/dev/test with a 20/5/90 document split. We are interested in the setting where only a small set of training data is available, to emulate use-cases in which annotating a large amount of data is impractical. We reserve more articles in the test set to ensure the evaluation is reliable.

\item \textbf{GICoref}~\cite{cao2020toward} consists of 95 documents from sources that include articles about non-binary people, fan-fiction from Archive of Our Own, and LGBTQ periodicals with a plethora of neopronouns (e.g., zie).
\item \textbf{MAP}~\cite{cao2020toward} consists of snippets of Wikipedia articles with two or more people and at least one pronoun.
\end{itemize}
We followed \citet{cao2020toward} to use LEA~\cite{moosavi2016which} as the evaluation metric for coreference clusters.

\subsection{Results on Violent Death Narratives}

We created 3 rules based on the approach described in Sec. \ref{sec:method}: (R1) Replace gendered terms with another gender. (R2) Replace gendered relationship terms with non-gendered terms. (R3) Replace terms describing gender with non-gendered terms.
Examples of the generated data are in Fig. \ref{fig:example2}.\footnote{See appendix for violent death narrative rules.}
When applying the augmented rules to the current 20/5 document split of the train/development (dev), we ended up with 100/25 train/dev documents enlarging both sets by 5 times. 

We compared the following models. 

\begin{compactitem}
\item \textbf{E2E\footnote{For all the models, the number of epochs of training is tuned on the development set.}} The E2E-Coref~\cite{lee2018higher-order} model trained on the OntoNotes 5.0 corpus. We used the implementation provided in the AllenNLP library~\cite{Gardner2017AllenNLP}. 
\item \textbf{E2E-FT} E2E-FT is a variant of E2E-Coref. It was trained on OntoNotes first and then fine-tuned on the 20 target training documents.
\item \textbf{E2E-Aug} E2E-Aug trained on OntoNotes first and then fine-tuned on the augmented target training documents.

\end{compactitem}

\begin{table}[t]
\centering
\small
    \begin{tabular}{  l  l  l l  p{1.1cm} } \toprule
     & Precision & Recall & F1 \\ \midrule
    E2E & 26.6 & 18.2 & 21.8 \\ 
     E2E-FT & \textbf{69.9} & 54.8 & 61.4 \\ 
    E2E-Aug & 68.7 & \textbf{57.9} & \textbf{62.8} \\ \bottomrule
    \end{tabular}
    \caption{Performance in LEA of each model on the Violent Death Narratives set. The E2E model trained on OntoNotes performs terribly on the VDN corpus due to the domain shift. With data augmentation, E2E-Aug significantly improves on the performance of E2E. }
    \label{table:LEA}
\end{table}

Results are shown in Table \ref{table:LEA}. By fine-tuning with a modest amount of in-domain data, E2E-FT significantly improved E2E in LEA F1. We saw E2E-Aug further improved E2E-FT by 5\% on LEA F1 with the 30 LGBT narratives in VDN's test set\footnote{Not found in tables}. Our results meaningfully improved the classification of LGBT-related data, and show the need for a more careful approach with data from underrepresented groups. Further, this improvement extended beyond our domain-specific data:  E2E-Aug further improved the E2E F1 score by 1.4\% in LEA F1 on the overall set. Overall, we saw a significant improvement when training coreference models with our augmented data, on both the overall and gender-neutral LGBT set.

\subsection{Results on GICoref and MAP}
We then evaluated the data augmentation approach on two publicly available datasets -- GICoref and MAP.
We experimented with the following 3 rules. (R4)  Randomly pick a person-cluster in the document and replace all pronouns in the cluster with a gender neutral pronoun (e.g., his $\leftarrow$ zir). (R5) Truncate the first name of each person. (R6) Same as the R4 but replacing only one pronoun in the cluster to the corresponding gender neutral pronoun. 

We followed \citet{zhao2018gender} and used GICoref and MAP only as the test data. 
We compared E2E with its variant E2E-Aug. The latter was trained on the union of the original dataset and variants of OntoNotes augmented using the above rules. 
We also compared our results with those from a E2E-Coref model trained on the union of the original and augmented data with the gender swapping rules described in \cite{zhao2018gender}. 



\begin{table}[t]
\small
     \begin{tabular}{@{} l l l p{1.1cm} @{}} \toprule
     & Precision & Recall & F1\\ \midrule
    E2E & 39.9 & 34.0 & 36.7 \\ 
    \citet{zhao2018gender} & 38.8 & 38.0 & 38.4 \\ 
    E2E-Aug-R4 & 40.5 & \textbf{43.8} & \textbf{42.1}  \\ 
    E2E-Aug-R5 & \textbf{41.2} & 41.1 & 41.2  \\ 
    E2E-Aug-R6 & 40.55 & 41.5 & 41.0  \\ 
    E2E-Aug-R456 & 40.7 & 41.9 & 41.3  \\ \bottomrule
    \end{tabular}
    \caption{Evaluation on GICoref. Results are in LEA. }
    \label{table:LEAGI}
\end{table}

\begin{table}[t]
\small
    \begin{tabular}{ @{} l  l  l @{} } \toprule
     & Binary-Pronouns & Neopronouns \\\midrule
    E2E & 36.6 & 24.3 \\
    E2E-Aug-R4 & \textbf{40.7} & \textbf{37.5} \\\bottomrule
    \end{tabular}
    \caption{Performance in LEA recall on binary-pronouns (male/female) and neopronouns clusters. }
    \label{table:NPVBP}
\end{table}

\begin{figure}[t]
\centering
\includegraphics[width=\linewidth]{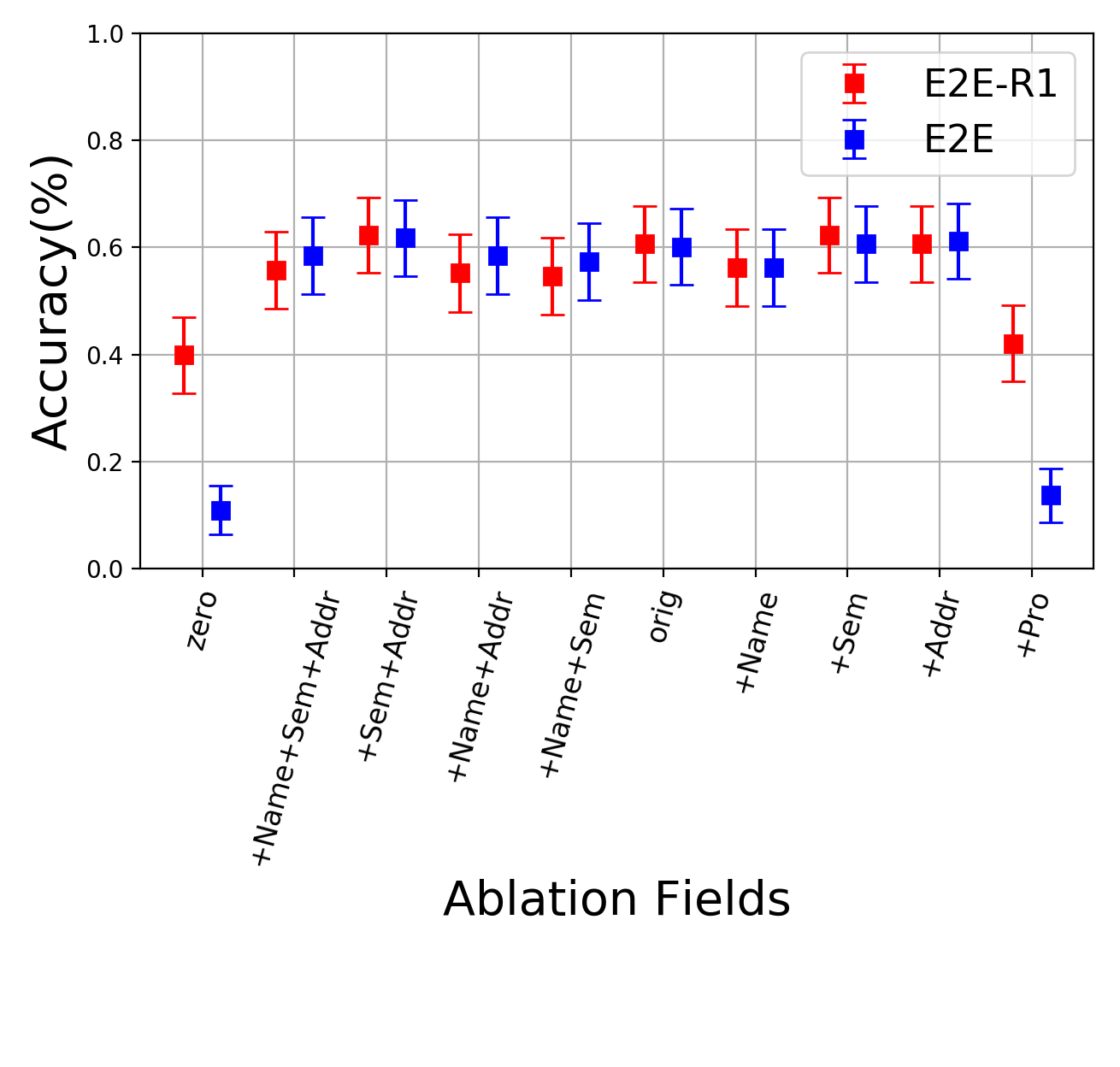}
\vspace{-25pt}
\caption{Performance in accuracy on the ablations of MAP. Error bar shows 95\% significance intervals.}
\label{fig:map}
\end{figure}

\paragraph{Results on GICoref} Results on GICoref are shown Table~\ref{table:LEAGI}. 
Few documents (0.3\%) in Ontonotes contained neopronouns.
Therefore, E2E struggled with resolving pronouns refering to LGBT individuals.  
\citet{zhao2018gender} had proposed to apply gender-swapping and entity anonymization to mitigate bias towards binary genders. However, their approach does not handle neopronouns and performs poorly compared to our models. 
In contrast, E2E-Aug improved E2E from a range of 4\% to 6\% in F1 with various data augmentation rules. 
When all the rules were applied, the performance was not superior to using only R4.



We further investigated the performance improvement of E2E-Aug-R4 on clusters containing binary pronouns and neopronouns.  
As shown in Table \ref{table:NPVBP}, E2E-Aug-R4 yielded a 4\% increase in recall among binary-gender pronouns and 12\% among neopronouns as compared with E2E. 
This reduced the performance gap between binary-gender pronouns and neopronouns from 12\% to 3\%. 
Our results show that R4 is highly effective, despite its simplicity.

\paragraph{Results on MAP} 
The core of MAP is constructed through different ablation mechanisms~\cite{cao2020toward}. Each ablation is a method for hiding various aspects of gender and then investigating the performance change of a model. Performance was evaluated based on the accuracy of pronoun resolution over the four label classes: person A, B, both, or neither.
We considered four ablation mechanisms as described in the Appendix. 
With these four possible ablations, each document was ablated a total nine times with each possible combination of ablations, producing a separate document. 

We compared E2E with E2E-R4 and showed the results in  Figure \ref{fig:map}.
E2E-R4 was better than or competitive with E2E in all the ablation scenarios. E2E-R4 especially outperformed E2E on the original set and the +Pro. set, where the performance was improved by 30\%. 


\section{Conclusion}
With policy decisions increasingly informed by computational analysis, it is imperative that methods used in these analyses be robust and accurate especially for marginalized groups. Our contributions improved coreference resolution for LGBT individuals, a historically underrepresented and marginalized population at high risk for suicide; they may improve the identification of LGBT individuals in NVDRS and hence inform better policy aimed to reduce LGBT deaths. More generally, we show how to use augmentation rules to adapt NLP models to real-world application domains where it is not feasible to obtain annotated data from crowdworkers. Finally, we introduced a novel dataset, VDN, which provide a challenging and consequential corpus for coreference resolution models. Our studies demonstrate the challenges of applying NLP techniques to real-world data involving diverse individuals (including LGBT individuals and their families) and suggest ways to make these methods more accurate and robust—thus contributing to algorithmic equity.

\section*{Discussion of Ethics}
Our research was exempted from human subjects review by the UCLA IRB. We applied for and were given access to the CDC's National Violent Death Reporting-System’s Restricted Access Database. As the data contain private information, we strictly follow their guidelines in our use of the dataset.

Despite our goal to improve gender inclusion in the coreference resolution system, we admit that our augmentation rules and data analyses may not fully address the diversities of sexual orientation in the population. Although our approach improves the performance of coreference systems, the final system is still not perfect and may exhibit some bias in its predictions.

\section*{Acknowledgements}
We thank all anonymous reviewers for their valuable feedback. We want to thank our three annotators Mika Baumgardner, Vivian Nguyen, and Mikaela Gareeb for their contributions to our paper. It is thanks to the amount of time they spent on annotations that we were able to produce this work. Our work for this study was partially funded by NIH( MH115344, MD006923). JGF was supported by an Infosys Membership in the School of Social Science at the Institute for Advanced Study.



\bibliography{naacl2021}

\begin{thebibliography}{19}
\expandafter\ifx\csname natexlab\endcsname\relax\def\natexlab#1{#1}\fi

\bibitem[{Babaeianjelodar et~al.(2020)Babaeianjelodar, Lorenz, Gordon,
  Matthews, and Freitag}]{10.1145/3366424.3383559}
Marzieh Babaeianjelodar, Stephen Lorenz, Josh Gordon, Jeanna Matthews, and Evan
  Freitag. 2020.
\newblock \href {https://doi.org/10.1145/3366424.3383559} {Quantifying gender
  bias in different corpora}.
\newblock In \emph{Companion Proceedings of the Web Conference 2020}, WWW '20,
  page 752–759, New York, NY, USA. Association for Computing Machinery.

\bibitem[{Cao and Daum{\'e}~III(2020)}]{cao2020toward}
Yang~Trista Cao and Hal Daum{\'e}~III. 2020.
\newblock Toward gender-inclusive coreference resolution.
\newblock In \emph{Proceedings of the 58th Annual Meeting of the Association
  for Computational Linguistics}, pages 4568--4595.

\bibitem[{Carnaz et~al.(2019)Carnaz, Quaresma, Nogueira, Antunes, and
  Ferreira}]{carnaz2019review}
Gon{\c{c}}alo Carnaz, Paulo Quaresma, Vitor~Beires Nogueira, M{\'a}rio Antunes,
  and Nuno NM~Fonseca Ferreira. 2019.
\newblock A review on relations extraction in police reports.
\newblock In \emph{World Conference on Information Systems and Technologies},
  pages 494--503. Springer.

\bibitem[{Clark et~al.(2020)Clark, Cochran, Maiolatesi, and
  Pachankis}]{pmid32453408}
K.~A. Clark, S.~D. Cochran, A.~J. Maiolatesi, and J.~E. Pachankis. 2020.
\newblock {{P}revalence of {B}ullying {A}mong {Y}outh {C}lassified as
  {L}{G}{B}{T}{Q} {W}ho {D}ied by {S}uicide as {R}eported in the {N}ational
  {V}iolent {D}eath {R}eporting {S}ystem, 2003-2017}.
\newblock \emph{JAMA Pediatr}.

\bibitem[{Gardner et~al.(2017)Gardner, Grus, Neumann, Tafjord, Dasigi, Liu,
  Peters, Schmitz, and Zettlemoyer}]{Gardner2017AllenNLP}
Matt Gardner, Joel Grus, Mark Neumann, Oyvind Tafjord, Pradeep Dasigi,
  Nelson~F. Liu, Matthew Peters, Michael Schmitz, and Luke~S. Zettlemoyer.
  2017.
\newblock \href {http://arxiv.org/abs/arXiv:1803.07640} {Allennlp: A deep
  semantic natural language processing platform}.

\bibitem[{Hovy et~al.(2006)Hovy, Marcus, Palmer, Ramshaw, and
  Weischedel}]{hovy2006ontonotes}
Eduard Hovy, Mitchell Marcus, Martha Palmer, Lance Ramshaw, and Ralph
  Weischedel. 2006.
\newblock {O}nto{N}otes: The 90{\%} solution.
\newblock In \emph{Proceedings of the Human Language Technology Conference of
  the {NAACL}, Companion Volume: Short Papers}, pages 57--60.

\bibitem[{Lee et~al.(2019)Lee, Yoon, Kim, Kim, Kim, So, and Kang}]{Lee_2019}
Jinhyuk Lee, Wonjin Yoon, Sungdong Kim, Donghyeon Kim, Sunkyu Kim, Chan~Ho So,
  and Jaewoo Kang. 2019.
\newblock \href {https://doi.org/10.1093/bioinformatics/btz682} {Biobert: a
  pre-trained biomedical language representation model for biomedical text
  mining}.
\newblock \emph{Bioinformatics}.

\bibitem[{Lee et~al.(2017)Lee, He, Lewis, and Zettlemoyer}]{lee2017end-to-end}
Kenton Lee, Luheng He, Mike Lewis, and Luke Zettlemoyer. 2017.
\newblock End-to-end neural coreference resolution.
\newblock In \emph{Proceedings of the 2017 Conference on Empirical Methods in
  Natural Language Processing}, pages 188--197.

\bibitem[{Lee et~al.(2018)Lee, He, and Zettlemoyer}]{lee2018higher-order}
Kenton Lee, Luheng He, and Luke Zettlemoyer. 2018.
\newblock Higher-order coreference resolution with coarse-to-fine inference.
\newblock In \emph{Proceedings of the 2018 Conference of the North {A}merican
  Chapter of the Association for Computational Linguistics: Human Language
  Technologies, Volume 2 (Short Papers)}, pages 687--692.

\bibitem[{Lira et~al.(2019)Lira, Xuan, Coleman, Swahn, Heeren, and
  Naimi}]{LIRA2019172}
Marlene~C. Lira, Ziming Xuan, Sharon~M. Coleman, Monica~H. Swahn, Timothy~C.
  Heeren, and Timothy~S. Naimi. 2019.
\newblock \href {https://doi.org/https://doi.org/10.1016/j.amepre.2019.02.027}
  {Alcohol policies and alcohol involvement in intimate partner homicide in the
  u.s.}
\newblock \emph{American Journal of Preventive Medicine}, 57(2):172 -- 179.

\bibitem[{Moosavi and Strube(2016)}]{moosavi2016which}
Nafise~Sadat Moosavi and Michael Strube. 2016.
\newblock Which coreference evaluation metric do you trust? a proposal for a
  link-based entity aware metric.
\newblock In \emph{Proceedings of the 54th Annual Meeting of the Association
  for Computational Linguistics (Volume 1: Long Papers)}, pages 632--642.

\bibitem[{Ng and Cardie(2002)}]{ng2002improving}
Vincent Ng and Claire Cardie. 2002.
\newblock Improving machine learning approaches to coreference resolution.
\newblock In \emph{Proceedings of the 40th Annual Meeting of the Association
  for Computational Linguistics}, pages 104--111.

\bibitem[{Ratner et~al.(2017)Ratner, Bach, Ehrenberg, Fries, Wu, and
  R{\'e}}]{ratner2017snorkel}
Alexander Ratner, Stephen~H Bach, Henry Ehrenberg, Jason Fries, Sen Wu, and
  Christopher R{\'e}. 2017.
\newblock Snorkel: Rapid training data creation with weak supervision.
\newblock In \emph{Proceedings of the VLDB Endowment. International Conference
  on Very Large Data Bases}, volume~11, page 269. NIH Public Access.

\bibitem[{Ream(2020)}]{REAM2020470}
Geoffrey~L. Ream. 2020.
\newblock \href
  {https://doi.org/https://doi.org/10.1016/j.jadohealth.2019.10.027} {An
  investigation of the lgbtq+ youth suicide disparity using national violent
  death reporting system narrative data}.
\newblock \emph{Journal of Adolescent Health}, 66(4):470 -- 477.

\bibitem[{Rudinger et~al.(2018)Rudinger, Naradowsky, Leonard, and
  Van~Durme}]{rudinger2018gender}
Rachel Rudinger, Jason Naradowsky, Brian Leonard, and Benjamin Van~Durme. 2018.
\newblock Gender bias in coreference resolution.
\newblock In \emph{Proceedings of the 2018 Conference of the North {A}merican
  Chapter of the Association for Computational Linguistics: Human Language
  Technologies, Volume 2 (Short Papers)}, pages 8--14.

\bibitem[{Soon et~al.(2001)Soon, Ng, and Lim}]{soon2001machine}
Wee~Meng Soon, Hwee~Tou Ng, and Daniel Chung~Yong Lim. 2001.
\newblock A machine learning approach to coreference resolution of noun
  phrases.
\newblock \emph{Computational Linguistics}, 27(4):521--544.

\bibitem[{Webster et~al.(2018)Webster, Recasens, Axelrod, and
  Baldridge}]{webster2018mind}
Kellie Webster, Marta Recasens, Vera Axelrod, and Jason Baldridge. 2018.
\newblock Mind the {GAP}: A balanced corpus of gendered ambiguous pronouns.
\newblock \emph{Transactions of the Association for Computational Linguistics},
  6:605--617.

\bibitem[{Zhao et~al.(2019)Zhao, Wang, Yatskar, Cotterell, Ordonez, and
  Chang}]{zhao2019gender}
Jieyu Zhao, Tianlu Wang, Mark Yatskar, Ryan Cotterell, Vicente Ordonez, and
  Kai-Wei Chang. 2019.
\newblock Gender bias in contextualized word embeddings.
\newblock In \emph{Proceedings of the 2019 Conference of the North {A}merican
  Chapter of the Association for Computational Linguistics: Human Language
  Technologies, Volume 1 (Long and Short Papers)}, pages 629--634.

\bibitem[{Zhao et~al.(2018)Zhao, Wang, Yatskar, Ordonez, and
  Chang}]{zhao2018gender}
Jieyu Zhao, Tianlu Wang, Mark Yatskar, Vicente Ordonez, and Kai-Wei Chang.
  2018.
\newblock Gender bias in coreference resolution: Evaluation and debiasing
  methods.
\newblock In \emph{Proceedings of the 2018 Conference of the North {A}merican
  Chapter of the Association for Computational Linguistics: Human Language
  Technologies, Volume 2 (Short Papers)}, pages 15--20.

\end{thebibliography}
\bibliographystyle{acl_natbib}
\newpage
\appendix
\section{Example Case for VDN with Augmenting Rules}
An example snippet of a case would be as follows:

... recently primary\_victim's \colorbox{yellow}{boyfriend} \colorbox{yellow}{$\rightarrow$ girlfriend} caught the primary\_victim cheating. primary\_victim's \colorbox{green}{boyfriend $\rightarrow$ partner} states that \colorbox{yellow}{he $\rightarrow$ she} and primary\_victim had a fight which got violent . primary\_victim is a 50 year old black \colorbox{red}{male $\rightarrow$ female}. ...

We see rule 1, rule 2, and rule 3 correlate to yellow, green, and red highlights. We applied each rule to the entire narrative.

\section{MAP ablations}
The core of MAP is constructed through different ablation mechanisms~\cite{cao2020toward}. Each ablation is a method to hiding various aspects of gender and investigate the performance change of a model.

\begin{enumerate}
  \itemsep0em 
  \item Replace third person pronouns with gender neutral variants (+Pro)
  \item Truncate the first name of each person in the document (+Name)
  \item Replace gendered nouns with the gender-neutral variant (+Sem)
  \item Remove terms of address (i.e. Mr., Mrs, etc.) (+Addr)
\end{enumerate}

In Figure 3, the ablations are applied individually and together, with zero containing all ablations. We see this yield 9 permutations, with the only ablation not being applied with others being +pro (except for zero).






\end{document}